\documentclass[11pt,a4paper]{article}
\usepackage[hyperref]{acl2020}
\usepackage{times}
\usepackage{latexsym}
\usepackage{graphicx}
\usepackage{amsmath}
\usepackage{makecell}
\usepackage{cjhebrew}

\usepackage{microtype}

\aclfinalcopy 

\title{Nakdan: Professional Hebrew Diacritizer}

\author{Avi Shmidman\textsuperscript{1} \qquad
  Shaltiel Shmidman\textsuperscript{1} \qquad
  Moshe Koppel\textsuperscript{1} \qquad
  Yoav Goldberg\textsuperscript{1,2} \\ 
 \textsuperscript{1}Bar Ilan University / Ramat Gan, Israel \\
DICTA / Jerusalem, Israel \\
\textsuperscript{2}Allen Institute for Artificial Intelligence \\
\texttt{avi.shmidman@biu.ac.il, \{shmidms1,koppel,yogo\}@cs.biu.ac.il}
}

\date{}

\begin{document}
\maketitle
\begin{abstract}
We present a system for automatic diacritization of Hebrew text.
The system combines modern neural models with carefully curated
declarative linguistic knowledge and comprehensive manually constructed tables
and dictionaries.
Besides providing state of the art diacritization accuracy, the system also
supports an interface for manual editing and correction of the automatic
output, and has several features which make it particularly useful for preparation of scientific editions of Hebrew texts. The system supports Modern Hebrew, Rabbinic Hebrew and Poetic Hebrew.
The system is freely accessible for all use at
\url{http://nakdanpro.dicta.org.il}.
\end{abstract}

\section{Introduction}

We present a web-based system for diacritization of Hebrew text, which caters to both casual and expert users. The diacritization engine driving the system combines manually curated 
linguistic resources with modern machine learning models.
\paragraph{Diacritization}
In Hebrew writing, the letters are almost entirely consonantal; the vowels are indicated by diacritic marks, generally positioned underneath the letters. However, in most cases, printed Hebrew omits the diacritic marks and includes only the letters, resulting in a highly ambiguous text, in which any given non-diacritized word can represent a host of different Hebrew words, each with a different meaning and pronunciation. For example, the form \<b.sl> can be diacritized as \<b*A.sAl> (noun, ``onion''), \<b*:.sel> (prefix+noun, ``in a shadow''), \<b*a.s*el> (prefix+definitive+noun, ``in the shadow'') and others.  The task of diacritization is thus a task of disambiguation: choosing from among the valid word possibilities for each non-diacritized word, and then adding in the diacritic marks accordingly. The multiple possibilities for diacritizing any given word often represent different morphological possibilities. Thus, to an extent, choosing the correct diacritization entails morphological disambiguation; conversely, prior morphological disambiguation greatly reduces the total possible diacritization possibilities.We provide further details in \S\ref{sec:diacritization}.

\paragraph{Hybrid Neural and Rule-based Approach}
Our approach, described in \S\ref{sec:approach}, uses several bi-LSTM-based 
deep-learning modules for disambiguating the correct diacritization in context.
However, it is also supplemented by comprehensive inflection tables and
lexicons, when appropriate. 

\begin{figure*}[t]
    \includegraphics[width=1\textwidth]{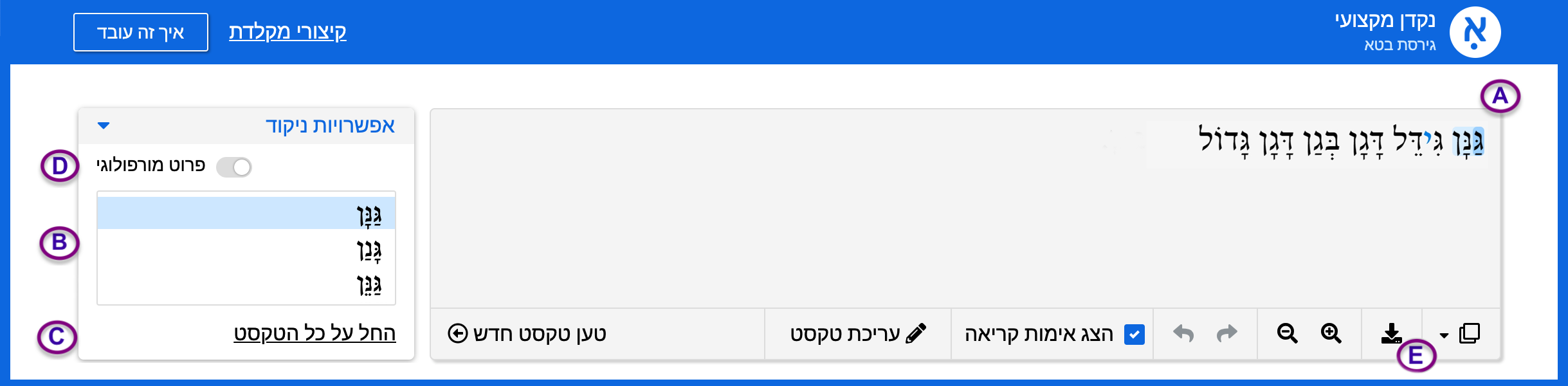}
    \caption{The main web interface of our diacritization tool, showing the automatic diacritized text (A) and allowing the user to proofread and potentially correct the text.  The user can navigate the words using the mouse or the left/right keys, and can select an alternate diacritization option from the listbox on the left (B) using either the mouse or the up/down keys. Changes for a given word can be marked for application over the entire text (C), and are marked in color (not shown in this example). The user can also choose to see the morphological analysis of each form (D). The resulting diacritized text can be exported to various formats (E).}
    \label{fig:main}
\end{figure*}

\paragraph{Web Interface}
We provide a web interface for the user to input a text for diacritization and refine the resulting diacritized text (Figure \ref{fig:main}).
Our system parses the text and automatically adds diacritics throughout. Afterward, the user can proofread the text in the interface. For each word, all alternate diacritization possibilities are provided for immediate selection,  ordered according to their predicted probability. Keyboard shortcuts allow efficient navigation of the text and fast selection of alternate options. Users can choose to see morphological analyses for each of the diacritization options, to assist in distinguishing between options. 

\paragraph{Diacritics in Scientific Editions}
We aim to provide a tool that is useful to casual users and language
enthusiasts, but also to experts and professionals who may use it to set
scientific editions of historical Hebrew texts. This latter requirement poses several challenges: handling of editorial sigla interspersed within the words; flexible handling of matres lectionis (letters which function as semi-vowels); and dealing with the orthography of medieval Hebrew, which often diverges widely from that of Modern Hebrew. Our tool meets scholarly requirements on all these fronts, as detailed in \S\ref{sec:advanced}.  

\section{The Hebrew Diacritics System}
\label{sec:diacritization}

The diacritics system of modern Hebrew marks vowels and gemination, and includes 12 primary diacritic symbols:
\begin{center}
\includegraphics[width=0.4\textwidth]{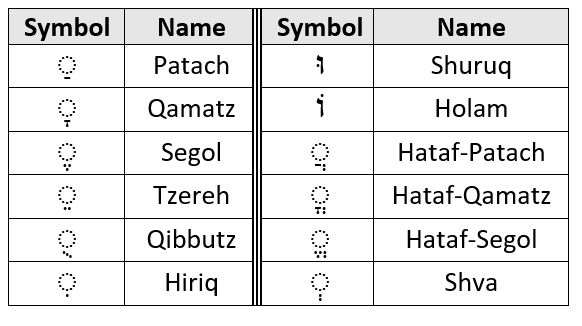}
\end{center}
Additionally, a dot in the middle of a letter indicates gemination. For the case of the 'shin' letter, an upper dot distinguishes between pronunciation as 's' or as 'sh'. 
Diacritized Hebrew aims to position a diacritic on every single letter of the word, with the exception of final letters and matres lectionis. 

\paragraph{Ambiguity}
In our tests, knowing the correct diacritics reduces the
full-morphological-analysis ambiguity from 9.1 to 2.4 average analyses per word form,
while knowing the full-morphological-analysis reduces the diacritization ambiguity 
from 6.2 to 1.4 average options per word form. Note that these numbers reflect fine-grained morphological tagging. If we utilize coarse-grained tagging, sufficing with the part of speech for each word, then knowing the correct diacritization reduces the average morphological ambiguity from 3.2 options to 1.97, while knowing the correct POS tag reduces the average diacritization ambiguity from 6.2 options to 2.75. Thus, the need for an automated diacritization utility is particularly crucial in order to properly disambiguate a Hebrew text.

\section{Approach}
\label{sec:approach}
Recent trends in NLP suggest moving towards machine-learned models that
automatically learn to extract the regularities in the data.
Such approaches have also been applied to diacritization of Arabic \cite{Belinkov2015ArabicDW, rashwan-arabic-diacritization, abandah-arabic-diacritization, mubarak-etal-2019-highly}.
However, while these generally provide
very strong results, they also often make mistakes that contradict our prior
knowledge of the linguistic system. While the machine-learned models generalize very well and
can learn to perform tasks in which humans cannot articulate the underlying
regularities, there are also many cases that language-experts \emph{can} articulate
precisely, and these tend to correlate with the cases that the learned models fail on.

We therefore take a hybrid approach. Similar to traditional diacritization
systems \cite{choueka1995nakdan}, we use our explicit knowledge about the language and the diacritization system whenever we can.
However, we also supplement our
knowledge with learned model predictions for the challenging cases for which we
cannot articulate the rules and regularities: selecting the appropriate
diacritization in context, and providing diacritization for out-of-vocabulary words.
This methodology departs from recent diacritization works that rely on HMM and
neural-network methods \cite{gal2002hmm,Belinkov2015ArabicDW}, while
ignoring forms of explicit linguistic knowledge.

We use such a combination of machine-learned and human-specified knowledge in
all the components of the system, either by supplementing the predictor with
manually constructed options, or by filtering its output space. 

Of course, a prerequisite for an effective machine-learned system is
high-quality training data. Our system is trained on a collection of 1,5M
diacritized tokens which we annotated in-house.

\section{High-quality Data Sources}
We make use of the following language resources and corpora, which we collected.
\paragraph{Language Resources}
\label{wordlists}
Our main resource is a high-coverage and accurate lexicon of Hebrew word forms,
their diacritization and their corresponding morphological analyses. Employing a staff of language experts, we began by assembling a list of all nouns, adjectives and verbal roots in the Hebrew language. This list includes 50K lexemes altogether (10K roots, 30,5K nouns, and 9,5K adjectives). We then built comprehensive inflection tables to generate all possible inflected forms from each of these lexemes, including all valid combinations of possessive and accusative suffixes, with full diacritization. Altogether, this process generated some 5,5 million inflected forms (3,8M verbal forms; 1,3M nominal forms; and 460K adjectival forms).
We also added 1,7K adverbs, and another 4,5K function words (conjunctions, prepositions, existentials, quantifiers, etc., including all possible suffix combinations). 
Finally, we collected a set of 17,5K frequent proper nouns (countries and major cities; heads of state and other notable people; and frequently-mentioned companies and organizations), and our language experts diacritized these as well.
These tables suffice for modern Hebrew; however, in historical Hebrew texts, we often find Aramaic terms interspersed within the Hebrew. Therefore, we also built a similarly comprehensive and diacritized wordlist for Babylonian Aramaic. Our Aramaic wordlist contains 750K verbal inflected forms; 200K nominals; 1,5K adjectives; and another 2K adverbs and function words.
We additionally assembled an exhaustive list of non-diacritized Hebrew names of persons
and locations (including collections of both street names and city names).

\paragraph{Annotated Corpora}
For morphological tagging, we make use of a corpus of 200K tokens of modern Hebrew, composed of
Hebrew fiction, news, wikipedia, and blogs. These tokens were manually annotated with fine-grained morphological information according to the scheme of \cite{elhadad2005hebrew}. Additionally, as noted, we annotated a 1,5M word diacritized modern Hebrew corpus, consisting of Hebrew prose (both fiction and non-fiction), newspapers (both news and op-ed), wikipedia, blogs (including many female-dominant blogs, to ensure coverage of feminine word forms), law protocols, Parliament proceedings, TV transcripts, academic texts, and biographical sketches. 
We have similarly collected and annotated corpora of historical Hebrew, consisting of Jewish legal writings and commentaries from the 3rd-12th centuries: 110K words with fine-grained morphological tagging, and 2M words with diacritization. Finally, regarding poetic Hebrew, we collected and annotated a corpus of 1,3M words, containing Hebrew poetry from both medieval and modern periods.

The undiacritized base texts were collected largely through partnerships with cooperating organizations in Israel; the morphological tagging and diacritization was done primarily in-house by our Hebrew language experts.

\section{System Architecture}

On a high level, our system works in the following stages, which we will
elaborate on below. Each stage combines engineered linguistic information and a
trained neural model.

\begin{enumerate}
        \item POS-tagging and morphological disambiguation.
        
        \item Filtering the possible diacritization analyses based on high coverage
        accurate tables and the output of stage (1).
        
        \item Ranking the possible diacritizations for each word, in context.
\end{enumerate}

\paragraph{Part-of-speech tagging and morphological disambiguation}
As diacritic marks closely interact with the morphological analysis and
part-of-speech (POS) of the token, we first perform POS-tagging and
morphological disambiguation, using a two-stage process. In the first stage,
each word is assigned its core part-of-speech, and in the second stage it is
enriched with additional morphological properties, where the set of
considered morphological properties is determined based on the coarse-grained
POS (e.g., nouns take gender, number and definiteness, while verbs do not take
definiteness but do take tense and person).\footnote{We consider the
following POS-tags: \textit{Adj, Adv, Conj, At\_Prep, Neg, Noun, Num, Prep, Pron, ProperNoun, Verb, Interrogative, Interj, Quantifier, Existential, Modal, Prefix, Participle, Copula, Titular, Shel\_Prep},
\normalfont
and the following morphological properties: \textit{Gender, Number, Person, Construct/Absolute, Suffix (possessive / accusative / pronominal)}.}

Training is performed on our annotated corpus of 200K tokens. The resulting tagger has an accuracy
of 92\% for the coarse-grained part-of-speech, and 79\% for full morphological
disambiguation.\footnote{While these numbers may seem low, we note that they are (a)
on-par with other Hebrew systems \cite{DBLP:conf/acl/AdlerE06,moretsarfatycoling2016} and (b) are only intended to support the diacritization process, where we find they do well.}

Both taggers are 2-layer bi-LSTM transducers \cite{goldberg-book}, where the
first stage coarse-grained tagger maps each token $w_i$ to a coarse POS-tag $t_i$, while the
second stage morphological tagger adds additional morphological properties
$m_i^1,...,m_i^k$. Each bi-LSTM takes as inputs vectors $x_1,...,x_n$
corresponding to tokens $w_1,...,w_n$ and produces vectors $h(x_1),...,h(x_n)$. These
vectors are then fed into multi-layer perceptrons (MLP) for predicting the POS-tags and
morphological properties, where each property is predicted by a different MLP:
\[
t_i = \arg\max_j \text{softmax}(MLP_{\text{pos}}(h(x_i))_{[j]}
\]
\[
m_i^k = \arg\max_j \text{softmax}(MLP_{m_k}(h(x_i))_{[j]}
\]

The set of MLPs $m_i^k$ for a word is determined based on its predicted coarse-grained
POS-tag.

In the coarse-grained tagger, each token $w_i$ is mapped to an input vector $x_i$ which encodes character
level information, distributional word-level information, possible morphological
analyses of $w_i$,\footnote{We find that providing the coarse-grained tagger with
information about possible fine-grained analyses of neighbouring words helps to disambiguate cases where a given word can be resolved as more than one POS. For instance, a given word may be resolvable as a noun or adjective; however, if the adjective possibility involves a feminine conjugation, and the preceding noun is a masculine noun, then the probability of the adjectival POS is severely reduced.} and lexicon-based features of $w_i$.
Specifically, $x_i$ is a concatenation of: (a) for a word $w_i$ made of
characters $c_1^{w_i},...,c_m^{w_i}$ the sum of
bi-LSTM states $\sum_j h(c_j^{w_i})$ from a char-level bi-LSTM that runs over the
entire sentence; (b) bi-LSTM state at $w_i$, for a word-level bi-LSTM that runs on pre-trained word2vec vectors for all of the words in the sentence; (c) a vector representing the possible fine-grained morphological analyses for the
word,\footnote{We assign trainable embeddings of 3-5 dimensions to each morphological category (gender, number, person, etc.), and we concatenate these together to form the input vector.} according to our wide-coverage lexicon; (d) bits indicating whether $w_i$
is in our comprehensive list of proper-nouns (names of streets, cities and people),
and whether it is in our wide-coverage lexicon at all (the latter is used to
mark rare and unknown words). In the fine-grained tagger, $x_i$ is a
concatenation of vector (b) above and: for a word $w_i$ where the predicted POS tag is $t_i$, and the possible fine-grained morphological analyses for $w_i$ limited by $t_i$ is represented by $m_i$, the bi-LSTM state for a bi-LSTM that runs on the concatenation of ($t_i$;$m_i$). Significantly, note that in the fine-grained tagger, $x_i$ does not include the information of the word form on the character-level. We find
this to be more accurate, because it removes  bias in cases where a specific character form happens to appear in the training corpus in only one configuration. This is particularly relevant regarding verbs which can be resolved as either a masculine or a feminine verb, each with a distinct diacritization. In many cases, the training corpus contains the verb only in one stereotypical gender configuration. By hiding the character-level information, we force the system to make a more logical morphological determination, because it is not able to mechanically set the feature equal to what was seen in the training corpus.

\paragraph{Constraints} The tagger predictions are constrained by a wide-coverage lexicon that maps word forms to their possible morphological analyses.
When a word is not in the lexicon, we allow all POS-tags for the word. We also
apply additional filters to rule out POS-tags for words that participate in a
hand-crafted list of about 10K word collocations, and in all of their possible inflected forms (e.g., in the context of the
tokens (\<byt mrq.ht>) \textit{byt mrk\^ht}, the word \<byt> \textit{byt} should not be tagged as the absolute form \<b*ayit> \textit{bayit}, but rather as the construct form \<b*eyt> \textit{beyt}. And thus too for the plural inflection of the same collocation - \<bty mrq.ht> \textit{bty mrk\^ht}, the word \<bty> \textit{bty} should not be tagged as \<b*it*iy> \textit{byty} (feminine noun with possessive suffix), but rather as the plural-construct form \<b*At*ey> \text{batey}).

\paragraph{Filtering}
For each word $w_i$ in the text, we retrieve from our wordlists (see
\S\ref{wordlists}) a set of possible diacritizations $D_i = d^i_1,...,d^i_\ell$ and their corresponding
morphological analyses. This set is then further refined by intersecting it with
the predicted morphological analysis for the word. Words that are not in our
list get an empty set, indicating that their diacritization is not
constrained. This stage leaves us with an average of 1.2 diacritic sequences
for each known word. If we were to perform random selection from this list, we
would achieve 87.1\% exact-match word-level diacritization accuracy on our Modern Hebrew test corpus. 

\paragraph{Diacritization Ranking}
Finally, we run an LSTM-based diacritization module to rank the possible
diacritization sequences from the previous stage, and to assign
diacritics to unknown words.

The LSTM-based module assigns a diacritic mark for each character in the
sequence.\footnote{Combinations of gemination with an additional diacritic mark are considered distinct diacritic symbols for prediction. An independent MLP predicts the position of the upper dot for the 'shin' character.}
The diacritics for each word $w_i$ are predicted separately, using beam-search over
the predictions of the diacritic for each letter with the word, to ensure word-level consistency. For known words, the
beam-search is constrained to valid diacritic predictions from the set $D_i$,
while for unknown words it is unconstrained.
Note that when predicting the diacritics for a letter $c^{w_i}_j$ in token $w_i$ the model is
aware of the other diacritic assignments in that word, but not of diacritic
assignments for the other words of the sentence. However, the model \emph{is}
context-aware, as it considers the character-level and word-level information
from the entire sentence via a sentence-level bi-LSTM layer.

To be more precise, each letter $c^{w_i}_j$ is mapped to a vector $h'(c^{w_i}_j)$ which is a concatenation of the followings two items: (a) bi-LSTM state at $c^{w_i}_j$ for a char-level bi-LSTM that runs over the entire sentence; (b) bi-LSTM state at $w_i$ for a word-level bi-LSTM that runs on the pre-trained word2vec vectors for all words in the sentence.
Then, for a given word $w_i$ we have a list of vectors representing each letter $h'(c^{w_i}_1)...h'(c^{w_i}_m)$. We then predict the diacritization sequence as follows. If this is a known word, then we have a list of $k$ possible diacritization sequences, and we choose the one with the highest score:
\[
s=\arg\max_k score(c^{w_i}_{1:m},t^{k}_{1:m}) \\[-2mm]
\]
where $t^k_{1:m}$ is the $k$th diacritic sequence, and $score(c^{w_i}_{1:m},t^{k}_{1:m})$ is calculated as:
\[
\sum_{j=1}^m MLP(h'(c_j^{w_i});LSTM(t^k_{0:j-1}))_{[t_j]} \\[-2mm]
\]

For unknown words, we run beam-search with $k=8$ to predict the $k$ most likely diacritization sequences, and we choose the top beam-ray. 

\section{Evaluation}

We evaluate the system quantitatively against two commercial Hebrew diacritization
systems, Morfix\footnote{\url{https://nakdan.morfix.co.il/}} and Snopi\footnote{\url{http://www.nakdan.com/}}, considered state-of-the-art.

We also provide qualitative evaluation, demonstrating the ability to diacritize unknown words, and to produce context-sensitive
diacritization.

\paragraph{Quantitative Evaluation}

We use two quantitative measures to evaluate our model. (1) Word-level accuracy: for a given word\footnote{For this calculation, punctuation and non-Hebrew words or symbols are ignored.}, we consider the prediction correct if and only if all the diacritic marks on the word are correct, including gemination and the 'shin' dot, with all matres lectionis removed. (2) Character-level accuracy: For each Hebrew letter in the input text we check if the model predicted the correct set of diacritic marks for the letter (and, for matres lectionis, we check that the model predicted their removal).

We evaluated the system on a 6,000-word unseen gold-test corpus, manually diacritized by a professional linguist  (Table \ref{tab:eval-results}). The corpus consists of a random selection of Hebrew wiki articles. We have made the test corpus publicly available.\footnote{The test corpus can be downloaded at this link: \url{http://tiny.cc/hebrew-test-git}}
\begin{table}[]
\centering
\scalebox{0.8}{
\begin{tabular}{|c|c|c|}
\hline
\multicolumn{1}{|l|}{} & \textbf{Letter Accuracy} & \textbf{Word Accuracy} \\ \hline
\textbf{Dicta}         & \textbf{95.12\%}         & \textbf{88.23\%}       \\ \hline
\textbf{Morfix}        & 90.32\%                  & 80.9\%                \\ \hline
\textbf{Snopi}         & 78.96\%                  & 66.41\%                 \\ \hline
\end{tabular}}
\caption{Accuracy on Modern Hebrew Test Corpus}
\label{tab:eval-results}
\end{table}

\begin{figure}[h]
    \includegraphics{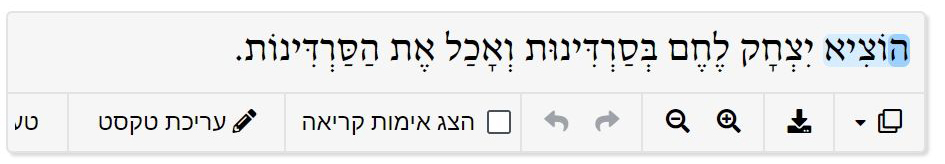}
    \caption{Diacritization of the fictional word \<srdynwt> in two different contexts, with two different prefixes; the word is diacritized as expected in both contexts.}
    \label{fig:sardinot}
\end{figure}

\paragraph{Qualitative Evaluation}
For the qualitative evaluation, we demonstrate that the system knows how to handle diacritization for unknown words, and this, in a context-sensitive manner.
For this example we choose an invalid word which conforms to Hebrew letter patterns but which does not actually exist in modern Hebrew:
\<srdynwt>. No such word exists in Hebrew dictionaries, nor in our wordlist. 
We put the word into a sentence in two contexts - in the first, it fills the role of an adverb, and in the second, it fills the role of a noun. Hebrew diacritization norms would dictate two different diacritizations for these two usages: for the adverb, the final vowel should be \textit{'u'}, while for the noun, it should be \textit{'o'}. Our system handles both correctly (Figure \ref{fig:sardinot}).

\section{Additional Text Genres}
In addition to modern Hebrew, we also support Rabbinic Hebrew and poetic Hebrew. These genres require specialized handling. Firstly, we cannot use our modern Hebrew morphology model, because the morphological and syntactic norms of these genres differ from those of modern Hebrew. Secondly, we cannot use our modern Hebrew wordlist filters. There is no standardized orthography for Rabbinic Hebrew, nor for medieval poetic Hebrew. Additionally, poets often specifically choose less common words in order to meet prosodic constraints; thus, our rare-word filters are not relevant. Finally, many words which would be considered invalid in modern Hebrew are found within these  other genres. Rabbinic Hebrew includes many Aramaic words, as well as Hebrew words with Aramaic prefixes. Poetic Hebrew includes oddities such as past-tense verbs with temporal prefixes. 

For Rabbinic Hebrew, we train a specialized morphology model based on our tagged historical Hebrew corpus. For poetry, where morphological sequences are less constrained and less predictable, we skip the morphology layer and diacritize the text directly based on the diacritization LSTM. 

In order to test our performance, we created test corpora for each of the genres. The poetry test corpus includes a set of liturgical poems of the 'yotzer' genre, transcribed from Cairo Genizah manuscripts.\footnote{Full data on these texts is available here: \url{http://weekdayyotzrot.com}} The Rabbinic Hebrew test corpus is taken from the 'Bet Yosef', a 16th century commentary on Jewish law.\footnote{Both test corpora are available for download here: \url{http://tiny.cc/hebrew-test-git}} In Tables \ref{tab:rab-results} and \ref{tab:poetry-results} we display our quantitative results on these two corpora. 

\begin{table}[]
\centering
\scalebox{0.8}{
\begin{tabular}{|c|c|c|}
\hline
\multicolumn{1}{|l|}{} & \textbf{Letter Accuracy} & \textbf{Word Accuracy} \\ \hline
\textbf{Dicta}         & \textbf{94.94\%}         & \textbf{87.94\%}        \\ \hline
\textbf{Morfix}        & 80.25\%                  & 68.1\%                 \\ \hline
\textbf{Snopi}         & 72.53\%                  & 58.39\%                \\ \hline
\end{tabular}}
\caption{Accuracy on Rabbinic Hebrew Test Corpus}
\label{tab:rab-results}
\end{table}

\begin{table}[]
\centering
\scalebox{0.8}{
\begin{tabular}{|c|c|c|}
\hline
\multicolumn{1}{|l|}{} & \textbf{Letter Accuracy} & \textbf{Word Accuracy} \\ \hline
\textbf{Dicta}         & \textbf{85.76\%}         & \textbf{70.23\%}       \\ \hline
\textbf{Morfix}        & 80.9\%                   & 65.3\%                 \\ \hline
\textbf{Snopi}         & 69.24\%                  & 52\%                   \\ \hline
\end{tabular}}
\caption{Accuracy on Poetic Hebrew Test Corpus}
\label{tab:poetry-results}
\end{table}

\section{Advanced Features}
\label{sec:advanced}

{\bf 1.} Scientific Editions: In scientific editions, editorial sigla are interspersed throughout the text. For instance, letters which are rubbed out in the textual witnesses will be supplied within brackets (\</s>[\<r>]\<md>). Existing diacritization tools fail here because they parse such sigla as word separators. Secondly, normative Hebrew diacritization entails the omission of matres lectionis, and indeed existing tools omit these letters when returning the diacritized text. However, in scientific editions, matres lectionis must be maintained in order to represent the manuscript evidence. Finally, the orthography of medieval Hebrew manuscripts can diverge wildly from modern norms; for example, we often find a yod inserted after the initial letter of a hitpael construction (e.g. \<hytlb/s>), a phenomenon which would never occur in a modern Hebrew text. Our tool meets all of these needs, and allows the user to either remove or maintain matres lectionis.

\begin{figure}[h]
\begin{center}
    \includegraphics[width=0.5\textwidth]{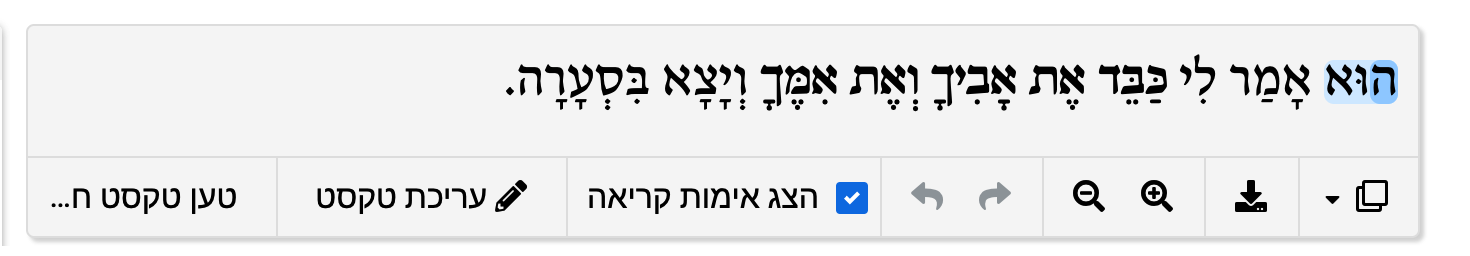}
\end{center}
    \caption{Integrated Biblical quote marked with font.}
    \label{fig:advanced-features-2}
\end{figure}

\noindent{\bf 2.} The web interface automatically highlights Biblical quotes within the Hebrew text. Biblical phrases are often incorporated into Hebrew texts, whether as explicit prooftexts or as rhetorical flourishes. We automatically identify such quotes, diacritize them according to the canonized diacritization of the Hebrew Bible, and display them in the distinctive Koren font (a font well-known for its use in modern Hebrew Bibles). See figure \ref{fig:advanced-features-2} for an example.

\section{Conclusion}
We are pleased to release our Hebrew diacritization system for free unrestricted use. It is powered by a combination of advanced machine learning and manually curated linguistic resources, and thus succeeds in setting a new state of the art for Hebrew diacritization. We have released also our diacritized test corpora for benchmarking. 
\bibliography{anthology,acl2020}
\bibliographystyle{acl_natbib}

\end{document}